\documentclass{article} % For LaTeX2
\usepackage{algorithm}
\usepackage{algorithmic}
\usepackage{booktabs}
\usepackage{iclr2024_conference,times}

\usepackage{amsmath,amsfonts,bm}

\def\eqref#1{equation~\ref{#1}}
\def\1{\bm{1}}

\DeclareMathAlphabet{\mathsfit}{\encodingdefault}{\sfdefault}{m}{sl}
\SetMathAlphabet{\mathsfit}{bold}{\encodingdefault}{\sfdefault}{bx}{n}

\usepackage{hyperref}
\usepackage{url}
\usepackage{float}
\usepackage{graphicx}

\title{Video-Teller: Enhancing Cross-Modal Generation with Fusion and Decoupling}
\author{
Haogeng Liu\textsuperscript{1,2}, Qihang Fan\textsuperscript{1,2}, Tingkai Liu\textsuperscript{3}, Linjie Yang\textsuperscript{3}, Yunzhe Tao\textsuperscript{3}, Huaibo Huang\textsuperscript{1}
\and
\textbf{Ran He\textsuperscript{1}}, \textbf{Hongxia Yang\textsuperscript{3}}\\
\textsuperscript{1}MAIS \& CRIPAC, Institute of Automation, Chinese Academy of Sciences, China.\\
\textsuperscript{2}School of Artificial Intelligence, University of Chinese Academy of Sciences, Beijing, China.\\
\textsuperscript{3}ByteDance, Inc.\\
\texttt{\{haogeng.liu,huaibo.huang\}@cripac.ia.ac.cn}\\
\texttt{fanqihang.159@gmail.com,rhe@nlpr.ia.ac.cn}\\
\texttt{\{tingkai.liu,yunzhe.tao,linjie.yang,hx.yang\}@bytedance.com}
}

\iclrfinalcopy % Uncomment for camera-ready version, but NOT for submission.
\begin{document}

\maketitle
\begin{abstract}
This paper proposes Video-Teller, a video-language foundation model that leverages multi-modal fusion and fine-grained modality alignment to significantly enhance the video-to-text generation task. Video-Teller boosts the training efficiency by utilizing frozen pretrained vision and language modules. It capitalizes on the robust linguistic capabilities of large language models, enabling the generation of both concise and elaborate video descriptions. To effectively integrate visual and auditory information, Video-Teller builds upon the image-based BLIP-2 model and introduces a \textit{cascaded Q-Former} which fuses information across frames and ASR texts. To better guide video summarization, we introduce a fine-grained modality alignment objective, where the cascaded Q-Former's output embedding is trained to align with the caption/summary embedding created by a pretrained text auto-encoder. Experimental results demonstrate the efficacy of our proposed video-language foundation model in accurately comprehending videos and generating coherent and precise language descriptions. It is worth noting that the fine-grained alignment enhances the model's capabilities ($4\%$ improvement of CIDEr score on MSR-VTT) with only 13\% extra  parameters in training and zero additional cost in inference.
%compared with the counterpart without fine-grained modality alignment.
\end{abstract}

\section{Introduction}

Large language models (LLMs) have made significant advancements~\citep{openai2023gpt4, chowdhery2022palm, bai2022constitutional}, and have subsequently been extensively utilized in multimodal tasks such as image-to-text generation and video-to-text generation~\citep{zhang2023transfer, mplug2, huang2023language, alayrac2022flamingo, wang2022ofa}, giving rise to a new class of models called multimodal large language model (MLLM). Prior to LLMs, video understanding models have been limited by the complexity of generated textual descriptions~\citep{videococa}, and downstream video-to-text tasks are constrained to short-form generations such as single-sentence video captioning. With the incorporation of large language models, models such as Video-LLaMA, VideoChat and Video-ChatGPT~\citep{videollama,videochat,maaz2023videochatgpt} are now capable of not only generating longer and nuanced video digests but also engaging in conversations grounded in video content. 

To leverage the power of pretrained LLMs such as LLaMA 2~\citep{touvron2023llama} and Vicuna~\citep{vicuna2023} without incurring the forbidding cost of retraining LLMs, MLLMs such as BLIP-2~\citep{blip2} have been proposed to integrate a trainable light-weight visual backbone with a frozen LLM via adaptor-like mechansims (such as the Q-Former proposed in BLIP-2). The expansion of BLIP-2 into the realm of video has quickly given rise to models such as Video-LLaMA~\citep{videollama}, which incorporates both visual and auditory information by prompting frozen LLM with embeddings computing by two corresponding encoders.
By relying on LLMs for modality integration, however, increases the computational cost during inference. Additionally, prior knowledge embedded in very large language models may negatively bias the generated video descriptions, leading to hallucination. Consequently, the issue of enhancing the accuracy of text generation in visual-language models while reducing the computational expense incurred has now become imminent.

\begin{figure*}[!t]
    \centering
    \includegraphics[width=0.85\linewidth]{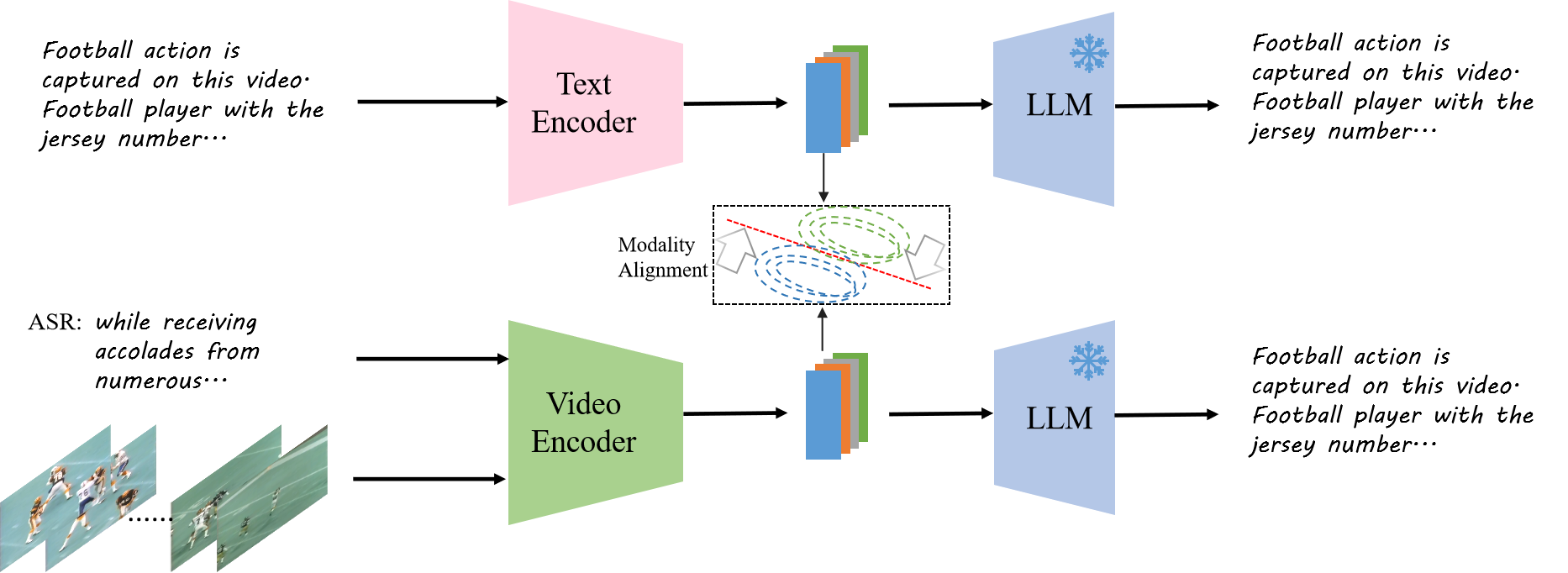}
    \caption{Overview of the proposed method. Here we show the detailed description generation (long-form text).}
    \label{fig:submodel}
\end{figure*}

In order to maintain the efficiency of model training while reducing computational overhead during inference, we propose cascaded Q-Former which fuses multi-frame visual information with auditory information prior to prompting LLMs, effectively reducing the computational overhead of LLM by half. Additionally, in contrast to Video-LLaMA's direct usage of raw audio, we leverage ASR information from videos as the representation for the audio modality to further enhance the model's comprehension capabilities. Due to the incorporation of additional modal information, to decouple crucial comments directly using methods similar to BLIP-2 becomes more difficult. Therefore, we propose the utilization of fine-grained modality alignment, thus enhancing the precision of content generated by the model and alleviating the issue of hallucination, we propose the utilization of fine-grained modality alignment as an auxiliary training approach. Figure \ref{fig:submodel} shows a high-level overview of the key concepts of Video-Teller.

We evaluated Video-Teller in both video-to-text generation and text-to-video retrieval tasks. Specifically, in video captioning, our approach achieves better results than the existing methods, such as HiTeA~\citep{ye2022hitea} with a smaller amount of data, thus confirming the effectiveness of fine-grained alignment and the integration of ASR. Additionally, in the task of long-text generation (video summarization), we obtained better performance (measured by BLEURT~\citep{sellam2020bleurt}) than the baseline model with larger frozen LLMs such as Video-LLaMA and VideoChat~\citep{videollama, videochat}. 

Overall, the main contributions of this paper are as follow.
\begin{itemize}
\item We propose Video-Teller, a video-language foundation model that integrates both the visual and speech information. Video-Teller reduces the computational cost of modality fusion by incorporating visual and ASR information through a cascaded Q-Former before the LLM.
\item We enhance video language learning by employing a text auto-encoder with LLM as the decoder to decouple textual features, enabling fine-grained alignment of video-text representations in an unsupervised manner. This approach improves the fusion of cross-modal information thus boosts the model's generation capability.
\item In addition to providing demonstrations of Video-Teller output, we quantitatively compare our proposed method with two representative MLLMs, Video-LLaMA~\citep{videollama} and VideoChat~\citep{li2023videochat}.
\end{itemize}

\section{Related work}
The pursuit foundation models that integrate and understand multiple modalities like vision and text have received enormous impetus from the research community in recently years. Previously, foundation models were largely end-to-end trainable models with architectures like dual-encoders~\citep{ALIGN, GLIP, multigrained, vlmo}, fusion-encoders~\citep{VisualBERT, MoCo, uniter, VL-BERT, oscar, vilbert, lxmert}, and encoder-decoders~\citep{coca, videococa}. However, these foundation models typically require fine-tuning of the entire model during adaptation, resulting in significant computational expenses. The advent of BLIP-2~\citep{blip2} changed the multimodal landscape by introducing a lightweight adaptor module, the Q-Former, which utilizes learnable queries to facilitate alignment of multiple modalities, reducing the need for fine-tuning the pre-trained language/visual models. Adaptor like modules have recently been remarkably successful in allowing multimodal researchers to tap into the tremendous natural language powers of large language models, with new models emerging every week such as InstructBLIP, VideoChat, Video-LLaMA and Mini-GPT4~\citep{instructblip, videochat, videollama, minigpt4}.

While adaptor modules like Q-Former can aid in merging input modalities, explicit alignments between modalities have traditionally been done via contrastive loss\citep{he2019moco} between a single textual \texttt{[CLS]} token and the other modalities. Specifically, previous models either align \texttt{[CLS]} tokens of text features and visual features for sample-level modal alignment~\citep{ALIGN, CLIP, coca}, or align \texttt{[CLS]} tokens with the rest of the tokens of the same modality using momentum~\citep{MoCo}. Such approaches enable modal alignment from a global perspective (via a single text token), but overlooks local information. 

In contrast, few foundation models have explored fine-grained alignment between tokens across modalities during pre-training, which we argue is crucial for detailed understanding of complex input data such as videos. ~\citep{li2022finegrained} propose LOUPE, which learns fine-grained semantic alignment from the novel perspective of game-theoretic interactions. While ~\citep{shukor2022efficient} leverage hierarchical cross-modal alignment loss for fine-grained modality alignment. These methods are highly effective, but they also tend to be more intricate. So in this paper, we propose to use a text auto-encoder to decouple the target text and use the decoupled feature to align with the video's hidden states.

\section{Method}

\subsection{Preliminaries: BLIP-2}
BLIP-2~\citep{blip2} is an image-text model designed to optimize training efficiency. It achieves this by utilizing pre-trained image encoders and frozen large language models. To address the modality gap, BLIP-2 introduces a lightweight Querying Transformer called Q-Former. Q-Former consists of learned queries and a transformer module with cross-attention. The input image initially undergoes the frozen Vision Transformer (ViT) to obtain an image patch token sequence. Subsequently, the learned queries interact with the image tokens through cross-attention. This process allows the input image to be encoded into a fixed-length sequence (as demonstrated in the paper, 32 tokens). These tokens are then projected and fed into the large language model to generate the corresponding text description of the image. Despite having $54\times$ fewer trainable parameters, BLIP-2 outperforms Flamingo80B~\citep{alayrac2022flamingo}. However, it should be noted that BLIP-2 is specifically designed to handle single-image inputs and cannot be directly applied to video-based applications.

\subsection{Model Architecture Of Video-Teller}
\begin{figure*}[!t]
    \centering
    \includegraphics[width=0.85\linewidth]{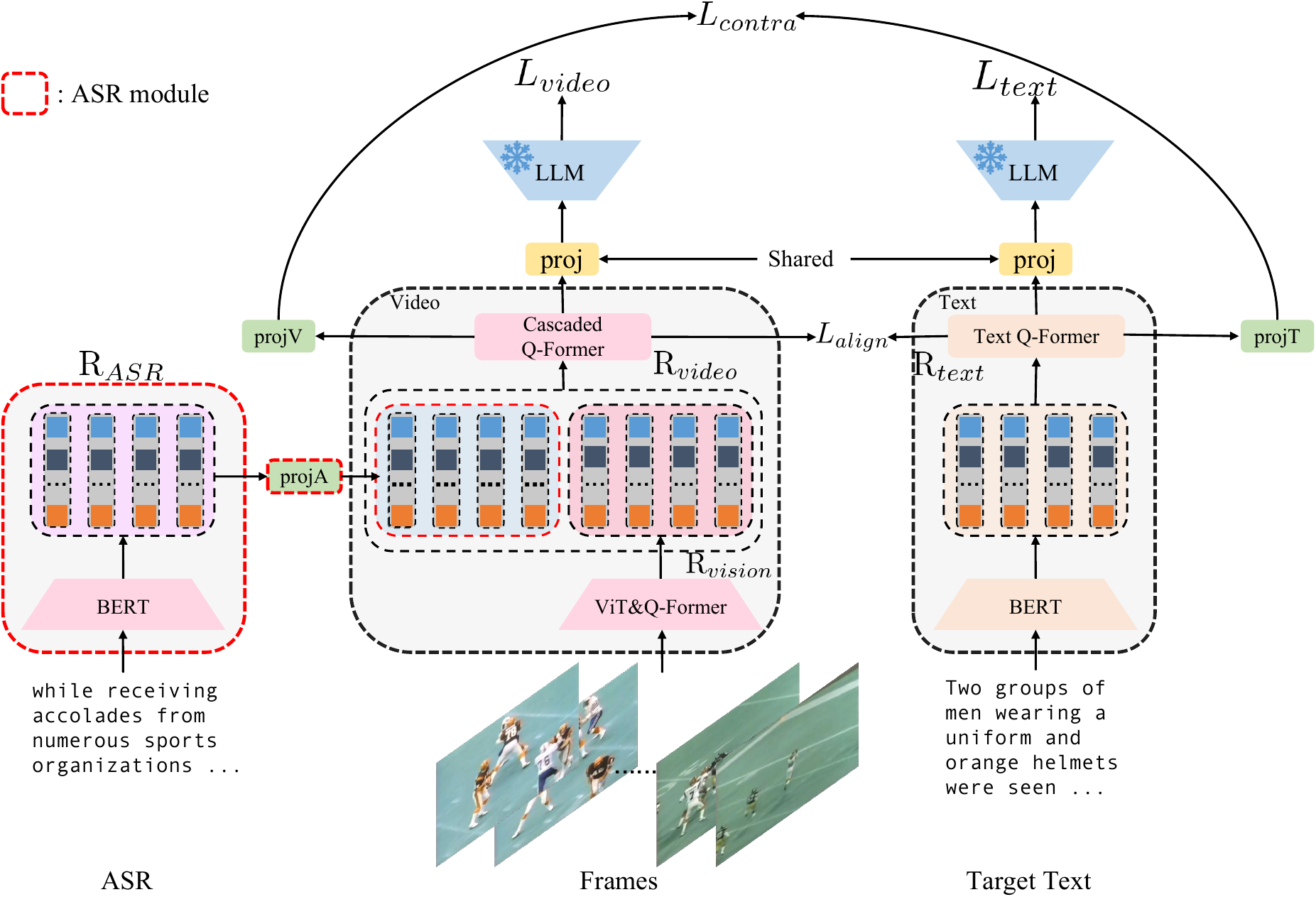}
    \caption{Overall architecture of the proposed model. The model consists of two primary branches. On the right-hand side, we have the text auto-encoder responsible for encoding the target text into a fixed-length representation denoted as $\mathbf{R}_{text}$. Conversely, on the left-hand side, we have the video module, which encodes the input video (comprising frames and ASR) into a video representation that shares the same shape as the text representation. Both of these representations are trained to reconstruct the target text by utilizing the LLM, while they are directly aligned through the Mean Squared Error (MSE) loss.}
    \label{fig:enter-label}
\end{figure*}

As illustrated in Figure \ref{fig:enter-label}, Video-Teller is composed of two primary components. The first component is a video foundation model, which takes frames and ASR texts as input and incorporates a LLM as the language decoder.
The second component is a text auto-encoder, which shares a similar structure with the video foundation model and also employs the same LLM as the language decoder. It is important to highlight that the text auto-encoder is exclusively utilized during the training phase and do not incur any additional computational cost during inference.

\subsubsection{Video-Teller}
The extension of BLIP-2 to process video input can be tackled via multiple approaches. One approach is to encode each frame individually via image-based BLIP-2 model, and prompt LLM directly using the concatenated frame-level embeddings. This approach, although straightforward and powerful, incurs considerable computational overhead as the input sequence length of the frozen LLM is now multiple by the number of frames sampled. Additionally, this approach relies entirely on LLM to perform modality integration and alignment. 

To address the aforementioned issue, we propose a cascaded Q-Former approach for integrating information from different video frames and texts generated by ASR. Let $\mathbf{V}\in \mathbb{R}^{F \times C \times H \times W}$ denote the input video frames, where $F$, $C$, $H$, and $W$ represent the number of frames, image channels, image height, and image weight, respectively. We utilize the vision encoder and Q-Former from BLIP-2 to individually extract the representation $\mathbf{R}_f \in \mathbb{R}^{Q_i \times E_i}$ for each frame, where $Q_i$ and $E_i$ indicate the number and size of query tokens in the image Q-Former. The aggregated visual features are obtained by concatenating all the image tokens from the Q-Former and are denoted as $\mathbf{R}{vision} \in \mathbb{R}^{F Q_i \times E_i}$.
%Then the whole features are flattened to obtain the video's visual representation $\mathbf{R}_{vision} \in \mathbb{R}^{(F \times Q_i) \times E_i}$. 

For ASR text, we first use encoder-only BERT \citep{devlin2019bert} to process it and obtain the encoded text features. We use the last hidden states of the text features $\mathbf{R}_{ASR} \in \mathbb{R}^E_i$ as the ASR tokens to be combined with the visual features. In order for the combined ASR and visual features to be consumed by the LLM, we need to further downscale the dimension of the combined features since the total number of tokens is too large to be directly handled by the LLM. Towards this end, we employed another transformer to reduce the number of tokens and to fuse the information from the ASR and visual features. We name this added transformer cascaded Q-Former and it adopts the same BERT structure as the original Q-Former with fixed number of query tokens to produce a fixed length of result tokens. We concat ASR tokens $\mathbf{R}_{ASR}$  with visual tokens $\mathbf{R}_{vision}$ as input to the cascaded Q-Former. Finally, we gain the representation of the whole video $\mathbf{R}_{video} \in \mathbb{R}^{Q_v \times E_v}$, where $Q_v, E_v$ denotes the query number and embedded dimension of cascaded Q-Former. Here we manually split $\mathbf{R}_{video}$ into two components, where the first component includes the first token of $\mathbf{R}_{video}$ that is used for video-text contrastive learning, and the second component contains the remaining 32 tokens that is used for fine-grained modality alignment and video-grounded text reconstruction.

\subsubsection{Text Auto-encoder}
We present a novel text auto-encoder that deviates from conventional models. Our approach leverages a frozen large language model as the decoder, while employing BERT and text Q-Former as the encoder to encode input text into a fixed-length representation (32 $\times$ 768). Our objective is to find the mapping between the input text and the soft prompt (fixed-length representation) that enables the LLM to recover the same input text. Our experiments show that, for one-sentence captions, it is sufficient to freeze the pretrained BERT module and only update weights of the text Q-Former.
However, multi-sentence summaries cannot be reconstructed with frozen BERT encoder. As a result, in our subsequent experiments, the entire text encoder (including both BERT and text Q-Former) is trainable.

\subsection{Fine-grained modality alignment}
As illustrated before, the text auto-encoder turns the input text into a fixed length intermediate representation, which covers the crucial information for text reconstruction. For video to text tasks, we aim at generating text with intermediate video representation. So we take the decoupled fixed length intermediate representation from text auto-encoder as intermediate target for video foundation model. This means we align video with corresponding text not only through the video-grounded text reconstruction but also the hidden feature's consistency, namely fine-grained modality alignment. Our proposed method is different from previous, instead of using game-theoretic~\citep{li2022finegrained} or hierarchical cross-modal alignment loss~\citep{shukor2022efficient} as we directly utilize the encoded tokens in our text auto-encoder as a learning target.

\section{Experiment}
\begin{figure}[!t]
    \centering
    \includegraphics[width=0.85\linewidth]{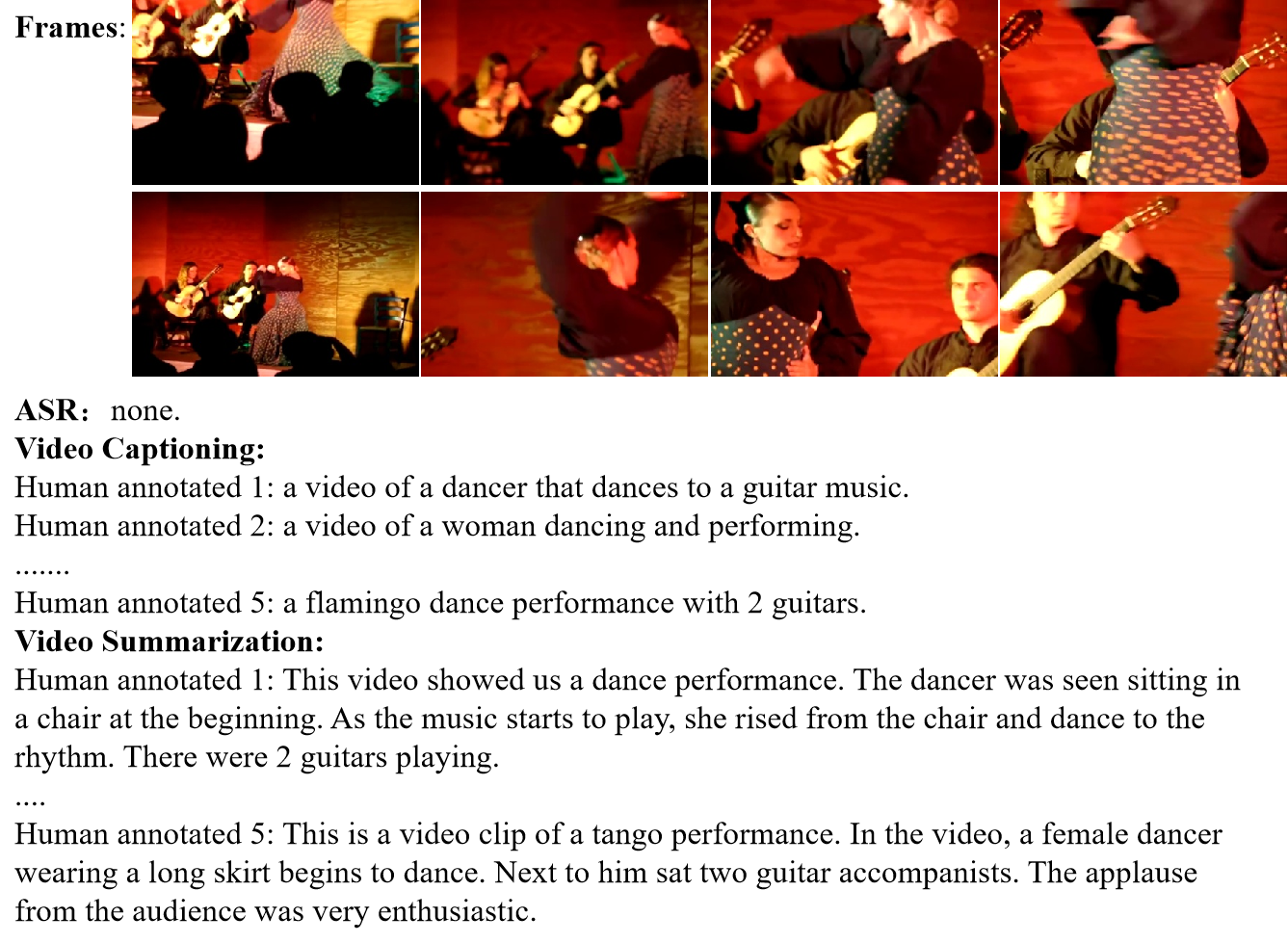}
    \caption{An example from Video-CSR. Here frames represent the video's vision information.}
    \label{fig:1}
\end{figure}
We evaluate our proposed method on several downstream tasks, including video captioning, video summarization and video retrieval.
\subsection{Setup}
\paragraph{Datasets}
We test our method's video understanding capability on MSR-VTT ~\citep{xu2016msr} and Video-CSR ~\citep{liu2023videocsr}. MSR-VTT is a large-scale video benchmark for video understanding, especially generating video captions. It covers 10K videos and each was annotated with about 20 natural sentences. There are 6513 videos for training and 2990 videos for testing. It should be noted that as MSR-VTT doesn't provide ASR, we use ``none.'' as the input to our ASR branch. Video-CSR is a newly released large-scale video benchmark for video understanding, covering roughly 5000 videos ranging from 15 seconds to 1 minute with each video annotated by human. There are 5 captions and 5 summaries for each video. Summaries are long captions that includes more details about the subject and activities in the video.  The average length of captions is 12.71 and the average length of summaries is 62.93 in Video-CSR. We adopt Video-CSR since its videos contain rich ASR information and is suitable to evaluate our framework with both visual and ASR input. Videos in this dataset can be divided into two parts, one part with rich ASR information while the other part with little ASR infromation. The ratio of videos with ample and limited ASR information is approximately 1 to 2. In cases that the video's ASR conveys only little information, we use the text ``none." as the ASR input.

For experiments on MSR-VTT, we use the WebVid-2M~\citep{bain2022frozen} and CC3M~\citep{sharma2018conceptual} (used as static video) for  pre-training. While in experiments on Video-CSR, we collect a pre-training dataset consists of 100K YouTube videos, where each video has 5 captions and 5 summaries generated by GPT-3.5 with the videos' metadata from YouTube, which covers description, ASR, comments and so on. The generated captions contains the key information of the video, but may miss some essential visual information if it is not described by the video's metadata. An example of a video from Video-CSR is shown in Figure \ref{fig:1}.

\vspace{-3mm}

\paragraph{Model Configurations}
We construct our model directly from the pre-trained BLIP-2, leveraging its extensive prior knowledge of images. For text auto-encoder, we use $\mathbf{BERT_{base}}$ to process the raw input. And then we use the first five layers of $\mathbf{BERT_{base}}$ as the text Q-Former. 33 learnable queries are used to project the text into the fix-length representation with cross-attention where the first one is for video-text contrastive learning. For the cascaded Q-Former, we construct it with the first 5 layers of the pre-trained $\mathbf{BERT_{base}}$ with 33 learnable query tokens. Empirically, we find that using more layers on the cascaded Q-Former and text Q-Former will deteriorate the performance. For ASR, the weights of $\mathbf{BERT}$ in its encoder is shared with the text auto-encoder. Totally, there are about 307M trainable parameters containing the image Q-Former, video Q-Former, text encoder, text Q-Former and a few linear projection layers. We apply pretrained opt-6.7b ~\citep{zhang2022opt} as our frozen LLM. For each video input, we sampled 8 frames as the vision representation. 

\vspace{-3mm}

\paragraph{Training and Evaluation}
We adopt a three-stage training process for our model. For the first stage, we pretrain the text auto-encoder. We use 32 Tesla-V100 GPUs, with a batch size of 8 on each individual GPU, and conducted training for two epochs. For the second stage, we train the whole model on video captioning or video summarization using 64 Tesla-V100 GPUs, with a batch size of 8 for video captioning and 6 for video summarization on each individual GPU. We train the model for another 2 epochs. For the third stage, we further finetune the model with 32 Tesla-V100 GPUs for 3 epochs on MSR-VTT and 10 epochs on Video-CSR.

\begin{table}[!t]
  \caption{Results for video caption on MSR-VTT. w/o A means without fine-grained modality alignment. SCST means Self-Critical Sequence Training~\citep{rennie2017selfcritical}. \textbf{Video-Teller achieves similar performance with its counterparts but uses much less PreTraining Data.}}
  \label{tb1}
  \centering
  \begin{tabular}{l|c|cccc}
    \toprule
    % & &\multicolumn{2}{c}{\textbf{Quality}}&  &\textbf{Speed }                \\
    Model  & \#PT Data   & B@4     & M & R &C \\
    \midrule
    HiTeA~\citep{ye2022hitea} &17M     & 49.2 & 30.7&65.0 &65.1    \\
    VideoCoCa~\citep{videococa} & 3B& 53.8  & -&68.0&73.2 \\
    GIT~\citep{wang2022git} &0.8B     & 53.8   & 32.9 &67.7&73.9 \\
    GIT2~\citep{wang2022git} &12.9B     & \textbf{54.8}   & 32.9 &\textbf{68.2}&\textbf{75.9} \\
    \midrule
    Video-Teller w/o A&4.5M     & 47.9  & 32.4 &65.5 &68.0 \\
    Video-Teller &4.5M     & 49.2   & 33.0 &66.4&72.0 \\
    Video-Teller (SCST)&4.5M     & 49.4   & \textbf{33.4} &67.0&74.5 \\

    \bottomrule
  \end{tabular}
\end{table}

\subsection{Video Captioning}
We conducted experiment on two datasets, MSR-VTT and Video-CSR. As Video-CSR is a newly released dataset, we implement baselines with  VideoCoCa~\citep{videococa} by adding a similar ASR fusion module to facilitate it to extract information from both frames and ASR text. We initialize VideoCoCa with CoCa pretrained on LAION-5B \citep{schuhmann2022laion5b}. We name the VideoCoCa model with the added ASR fusion module VideoCoCa (ASR). Results on MSR-VTT can be found in Table \ref{tb1} and results on Video-CSR can be found in Table \ref{tb2}. All results are reported on BLEU-4 (B@4), METEOR (M), CIDEr (C) and ROUGE-L (R). 

We also test the model applying self-critical sequence training, which is a REINFORCE algorithm that directly optimize the CIDEr metric~\citep{rennie2017selfcritical}. Those results demonstrate Video-Teller's strong video description though using limited videos for pre-training compared with other models.

\begin{table}[!t]
  \caption{Results for video caption on Video-CSR. w/o A means without fine-grained modality alignment. For Zero-Shot, both models are trained 100K videos from pretraining dataset.}
  \label{tb2}
  \centering
  \begin{tabular}{l|c|cccc|cccc}
    \toprule
    % & &\multicolumn{2}{c}{\textbf{Quality}}&  &\textbf{Speed }                \\
    & &\multicolumn{4}{c}{\textbf{Finetuned}}&\multicolumn{4}{|c}{\textbf{Zero-Shot}}   \\
    Model &ASR  & B@4     & M & R &C & B@4     & M & R &C \\
    \midrule
    VideoCoCa&No & 6.2  & 11.0&23.8&18.7 & 2.1  & 10.7&18.7&5.7 \\
    VideoCoCa (ASR)&Yes& 7.1   & 11.9 &25.0&22.1 & 2.8   & 11.4 &19.7&9.1 \\
    \midrule
    Video-Teller w/o A&Yes     &7.2  & 12.7 &26.3 &21.9 &3.5  & 12.2 &22.2 &13.9\\
    Video-Teller&Yes     & \textbf{10.4}   & \textbf{14.7} &\textbf{28.7}&\textbf{30.7} &\textbf{5.6}  & \textbf{14.2} &\textbf{24.0} &\textbf{19.9}\\
    \bottomrule
  \end{tabular}
\end{table}
\subsection{Video Summarization}
We evaluate performance of video summarization on Video-CSR. This dataset covers 5000 videos. We randomly choose 1200 videos for testing, while the rest are used for fine-tuning the models. It is important to mention that the ratio of videos with ample and limited ASR information in the test set and training set is both approximately 1 to 2. We compare with four baseline models: VideoCoCa, VideoCoCa (ASR), Video-LLaMA~\citep{videollama}, and VideoChat~\citep{li2023videochat}. Among them, VideoCoCa and Video-LLaMA only uses visual input and VideoCoCa (ASR) and VideoChat uses both visual and ASR input. We also evaluate both zero-shot and finetuned performance. For the metrics, we choose BLEURT \citep{sellam2020bleurt} as the main metrics. We also report results with CIDEr (C) and ROUGE-L (R). Results can be found in Table \ref{tb3}.  After calculating the metrics, we randomly select 20 generated sentences from different models. We manually ranked each result to assess their level of consistency with various indicators and find that semantic-related evaluation metrics such as BLEURT \citep{sellam2020bleurt} are more suitable than metrics based on string matching for long text evaluation. The results also indicate that Video-Teller has achieved certain advantages in video summarization compared to other models.
\begin{table}[!t]
  \caption{Results for video summarization on Video-CSR. w/o A means without fine-grained modality alignment.}
  \label{tb3}
  \centering
  \begin{tabular}{l|c|ccc|ccc}
    \toprule
    % & &\multicolumn{2}{c}{\textbf{Quality}}&  &\textbf{Speed }                \\
    & & \multicolumn{2}{r}{\textbf{Finetuned}}& &\multicolumn{2}{r}{\textbf{Zero-Shot}}   &  \\
    Model  & \#PT Data   &BLEURT & R &C &BLEURT & R &C\\
    \midrule
    VideoCoCa & 0.5M& 29.6&19.3&2.9 & 28.8&18.6&3.0 \\
    VideoCoCa (ASR) & 0.5M& 36.8&22.4&9.5 & 31.0&20.1&8.1 \\
    Video-LLaMA  & - & -&-&- & 39.3&19.2&2.1 \\
    VideoChat & - & - & - & - & 42.8 & 22.6 & 15.2\\
    \midrule
    Video-Teller w/o A&0.5M     &  45.2&22.4 &9.7 & 41.2&20.1&7.1 \\
    Video-Teller &0.5M     & \textbf{47.1 }&\textbf{23.5}&\textbf{11.2}& \textbf{43.3}&21.3&9.0\\
    \bottomrule
  \end{tabular}
\end{table}

\subsection{Ablation Experiments}
While our model exhibits commendable performance in the text generation task, we remain skeptical about the extent to which the inclusion of ASR and fine-grained alignment can genuinely enhance its performance. Consequently, we undertake ablation experiments and assess them using Video-CSR and MSR-VTT dataset. Results on Video-CSR can be found in Table \ref{tb4}. Results for MSR-VTT are in Table \ref{tb4}. As MSR-VTT doesn't provid ASR, we only test the influence of alignment and contrastive loss. From the results, we can observe that on Video-CSR our model's performance declines when either ASR or fine-grained alignment is absent. This demonstrates the effectiveness of our approach on real-world scenario datasets. Result on MSR-VTT captioning also shows the fine-grained alignment improves the performance. 
\begin{table}[!t]
  \caption{Results for video captioning. Here w/o A means without align while w/o C means without contrastive learning. We also use w/o ASR represents without ASR.}
  \label{tb4}
  \centering
  \begin{tabular}{l|cccc|cccc}
    \toprule
    % & &\multicolumn{2}{c}{\textbf{Quality}}&  &\textbf{Speed }                \\
    & &\multicolumn{2}{c}{\textbf{Finetuned}}& & &\multicolumn{2}{c}{\textbf{Zero-Shot}}   &  \\
    Model    & B@4     & M & R &C & B@4     & M & R &C \\
    \midrule
    Results on Video-CSR \\
    \midrule    
    Video-Teller w/o ASR   &4.7  & 9.8 &22.8 &13.1 &2.2 & 10.5 &19.7 &13.4\\
    Video-Teller w/o A    &7.2  & 12.7 &26.3 &21.9 &21.9  & 12.2 &22.2 &13.9\\
    Video-Teller w/o C    & 10.3   & 14.7 &28.5 &30.4 & 5.6 & 14.1 &24.0 &19.8\\
    Video-Teller    & \textbf{10.4}   & \textbf{14.7} &\textbf{28.7}&\textbf{30.7} &\textbf{5.6}  & \textbf{14.2} &\textbf{24.0} &\textbf{19.9}\\
    \midrule
    Results on MSR-VTT \\
    \midrule
    Video-Teller w/o A    &47.9  &31.5  &65.3 &69.6& 12.4 & 17.6 &36.3 &24.6\\
    Video-Teller w/o C    & 48.4   & 32.9 &65.7 &70.9&13.4  &18.7 &38.5&25.4\\
    Video-Teller    & \textbf{49.2}   & \textbf{33.0} &\textbf{66.4}&\textbf{72.0}&\textbf{15.6} & \textbf{19.6}&\textbf{40.1} &\textbf{26.9} \\
    \bottomrule
  \end{tabular}
\end{table}

\subsection{Video Retrieval results}
Though achieving strong results on video generation task with fine-grained modality alignment, it still needs to be verified whether the method will have an impact on the accuracy of retrieval. Through ablation experiments, it's demonstrated that fine-grained modality alignment enhances the cross-modal generation capability of the model without affecting its retrieval accuracy. Result can be found in Table \ref{tb5}. The model is pre-trained with WebVid-2M \citep{bain2022frozen} and CC3M \citep{sharma2018conceptual}.
\begin{table}[!t]
  \caption{Results for video retrieval on MSR-VTT, where w/o A means without fine-grained alignment.}
  \label{tb5}
  \centering
  \begin{tabular}{l|c|ccc|ccc}
    \toprule
    & &\multicolumn{3}{c}{\textbf{Finetuned}} &\multicolumn{3}{|c}{\textbf{Zero-Shot}}  \\
    Model  & \#PT Data   & R@1     & R@5 & R@10 & R@1     & R@5 & R@10  \\
    \midrule
    Video-Teller w/o A &4.5M & 33.1  & 57.8 & 65.9 & 41.0  & 67.1 & 77.3 \\
    Video-Teller &4.5M  & 33.5  & 57.5  & 66.1 & 40.7  & 67.5  & 77.7 \\
    \bottomrule
  \end{tabular}
\end{table}
From above result, we have demonstrated that fine-grained alignment can enhance the generation capability of the model without affecting video retrieval task.

\section{Analysis}
As shown before, we find that Video-Teller, with limited video data for pre-training, achieves strong performance both on video summarization and video captioning. We will analyze the improvements to the model that our proposed method brings in terms of hallucination of description.

Similar to LLM, Video-Teller is bothered by hallucination. it tends to fill in incorrect information, especially when generating detailed description. We evaluated the severity of different models' hallucination through manual assessment. Specifically, we randomly selected 50 generated results from the test set of Video-CSR and categorized them into three types: no hallucination, slightly hallucination, and severe hallucination, based on the comparison between the generated content and manually annotated content. The ratios for each model are illustrated in Table \ref{tb6}. We also provide criteria for rating different levels of hallucination in Table~\ref{tb level}.

\begin{table}[H]
  \caption{Criteria of rating hallucination.}
  \label{tb level}
  \centering
  \begin{tabular}{l|p{9cm}}
    \toprule
    Hallucination level  &  Description  \\
    \midrule
    no hallucination&The predicted summary delineates events that are entirely congruous with the actual video, albeit with potential omissions in its depiction.\\
    \midrule
    moderate hallucination &The predicted summary portrays events that are largely congruent with the actual video, albeit with some minor deviations in certain details.
\\
\midrule
    severe hallucination&The predicted summary depicts events that are starkly divergent from the actual video.
 \\
    \bottomrule
  \end{tabular}
\end{table}

\begin{table}[H]
  \caption{Hallucination ratio of different models. w/o A means trained without fine-grained alignment.}
  \label{tb6}
  \centering
  \begin{tabular}{l|ccc}
    \toprule
    % & &\multicolumn{2}{c}{\textbf{Quality}}&  &\textbf{Speed }                \\
    Model  &  no hallucination  & moderate hallucination  & severe hallucination\\
    \midrule
    VideoCoCa (ASR)&0.60 & 0.26 & 0.14\\
    Video-LLaMA &0.26 &0.40& 0.34\\
    Video-Teller w/o A& 0.40 & 0.26  & 0.34 \\
    Video-Teller & 0.56& 0.24   & 0.20\\

    \bottomrule
  \end{tabular}
\end{table}
Based on the findings presented in Table \ref{tb6}, it is evident that models utilizing LLMs face a more pronounced issue of hallucination. This can be attributed to the limited information provided by the visual encoder, forcing the LLM to heavily rely on imaginative processes to complete the description. In contrast, VideoCoCa, which does not employ an LLM, exhibits a relatively milder form of hallucination. This difference can be explained by VideoCoCa's tendency to generate shorter descriptions when faced with insufficient information, thereby reducing the generation of extraneous content. Conversely, the extensive prior knowledge of the LLM engenders the production of erroneous information. With our fine-grained alignment, Video-Teller is able to significantly reduce the rate of hallucination, with the no hallucination rate increased from 0.40 to 0.56, and severe hallucination rate diminished from 0.34 to 0.20. This indicates that the fine-grained alignment enforces the encoded video tokens $\mathbf{R}_{video}$ to be more relevant to the semantics of the target caption/summary and thus reduces hallucination. We provide a demo which shows model trained without fine-grained alignment suffers more from hallucination in Figure \ref{fig:2} in the appendix.

As we could see in Figure \ref{fig:2}, this video belongs to the category with a high ASR (Automatic Speech Recognition) content. Therefore, in order to generate its summary, it is necessary to make better use of the ASR information. From the ASR information, we can infer that this video discusses the relevant aspects of the decline in clean energy prices, just as predicted by Video-Teller. However, we can observe that without alignment, the model's description includes specific price changes that cannot be extracted from the video. 
%This indicates that it is an illusion of the model.

\section{Conclusion}
In this paper, we propose Video-Teller, a robust video-text foundation model that attains impressive performance on video-to-text tasks, encompassing both concise and comprehensive descriptions. Video-Teller leverages the rich speech information contained in the videos to enhance the model's understanding of the video. Simultaneously, it utilizes pre-trained visual models and large language models to reduce training costs while maintaining the impressive performance. Furthermore, we employ a standalone text auto-encoder to learn the proper intermediate language tokens that guides the learning of the video foundation model, which boosts the decoupling of the fused multi modality information. Extensive experimental results demonstrate the impressive performance of our approach with light-weighted training, effectively reducing model hallucinations (no hallucination rate wit a gain from 40\% to 56\%) and significantly improving the accuracy of model descriptions (BLEURT score increased from 41.2 to 43.3).

\bibliography{iclr2024_conference}

\begin{thebibliography}{45}
\providecommand{\natexlab}[1]{#1}
\providecommand{\url}[1]{\texttt{#1}}
\expandafter\ifx\csname urlstyle\endcsname\relax
  \providecommand{\doi}[1]{doi: #1}\else
  \providecommand{\doi}{doi: \begingroup \urlstyle{rm}\Url}\fi

\bibitem[Alayrac et~al.(2022)Alayrac, Donahue, Luc, Miech, Barr, Hasson, Lenc, Mensch, Millican, Reynolds, et~al.]{alayrac2022flamingo}
Jean-Baptiste Alayrac, Jeff Donahue, Pauline Luc, Antoine Miech, Iain Barr, Yana Hasson, Karel Lenc, Arthur Mensch, Katherine Millican, Malcolm Reynolds, et~al.
\newblock Flamingo: a visual language model for few-shot learning.
\newblock \emph{Advances in Neural Information Processing Systems}, 35:\penalty0 23716--23736, 2022.

\bibitem[Bai et~al.(2022)Bai, Kadavath, Kundu, Askell, Kernion, et~al.]{bai2022constitutional}
Yuntao Bai, Saurav Kadavath, Sandipan Kundu, Amanda Askell, Jackson Kernion, et~al.
\newblock Constitutional ai: Harmlessness from ai feedback, 2022.

\bibitem[Bain et~al.(2022)Bain, Nagrani, Varol, and Zisserman]{bain2022frozen}
Max Bain, Arsha Nagrani, Gül Varol, and Andrew Zisserman.
\newblock Frozen in time: A joint video and image encoder for end-to-end retrieval, 2022.

\bibitem[Bao et~al.(2022)Bao, Wang, Dong, Liu, Mohammed, Aggarwal, Som, Piao, and Wei]{vlmo}
Hangbo Bao, Wenhui Wang, Li~Dong, Qiang Liu, Owais~Khan Mohammed, Kriti Aggarwal, Subhojit Som, Songhao Piao, and Furu Wei.
\newblock {VLMo}: Unified vision-language pre-training with mixture-of-modality-experts.
\newblock In \emph{NeurIPS}, 2022.

\bibitem[Chen et~al.(2020)Chen, Li, Yu, Kholy, Ahmed, Gan, Cheng, and Liu]{uniter}
Yen-Chun Chen, Linjie Li, Licheng Yu, Ahmed~El Kholy, Faisal Ahmed, Zhe Gan, Yu~Cheng, and Jingjing Liu.
\newblock Uniter: Universal image-text representation learning.
\newblock In \emph{ECCV}, 2020.

\bibitem[Chiang et~al.(2023)Chiang, Li, Lin, et~al.]{vicuna2023}
Wei-Lin Chiang, Zhuohan Li, Zi~Lin, et~al.
\newblock Vicuna: An open-source chatbot impressing gpt-4 with 90\%* chatgpt quality, March 2023.

\bibitem[Chowdhery et~al.(2022)Chowdhery, Narang, Devlin, Bosma, et~al.]{chowdhery2022palm}
Aakanksha Chowdhery, Sharan Narang, Jacob Devlin, Maarten Bosma, et~al.
\newblock Palm: Scaling language modeling with pathways, 2022.

\bibitem[Dai et~al.(2023)Dai, Li, Li, Tiong, Zhao, Wang, Li, Fung, and Hoi]{instructblip}
Wenliang Dai, Junnan Li, Dongxu Li, Anthony Meng~Huat Tiong, Junqi Zhao, Weisheng Wang, Boyang Li, Pascale Fung, and Steven Hoi.
\newblock Instructblip: Towards general-purpose vision-language models with instruction tuning, 2023.

\bibitem[Devlin et~al.(2019)Devlin, Chang, Lee, and Toutanova]{devlin2019bert}
Jacob Devlin, Ming-Wei Chang, Kenton Lee, and Kristina Toutanova.
\newblock Bert: Pre-training of deep bidirectional transformers for language understanding, 2019.

\bibitem[He et~al.(2020)He, Fan, Wu, Xie, and Girshick]{he2019moco}
Kaiming He, Haoqi Fan, Yuxin Wu, Saining Xie, and Ross Girshick.
\newblock Momentum contrast for unsupervised visual representation learning.
\newblock 2020.

\bibitem[Huang et~al.(2023)Huang, Dong, Wang, Hao, et~al.]{huang2023language}
Shaohan Huang, Li~Dong, Wenhui Wang, Yaru Hao, et~al.
\newblock Language is not all you need: Aligning perception with language models, 2023.

\bibitem[Jia et~al.(2021)Jia, Yang, Xia, Chen, et~al.]{ALIGN}
Chao Jia, Yinfei Yang, Ye~Xia, Yi-Ting Chen, et~al.
\newblock Scaling up visual and vision-language representation learning with noisy text supervision.
\newblock In \emph{ICML}, 2021.

\bibitem[Li et~al.(2022)Li, He, Wei, Qian, Zhu, Xie, Zhuang, Tian, and Tang]{li2022finegrained}
Juncheng Li, Xin He, Longhui Wei, Long Qian, Linchao Zhu, Lingxi Xie, Yueting Zhuang, Qi~Tian, and Siliang Tang.
\newblock Fine-grained semantically aligned vision-language pre-training, 2022.

\bibitem[Li et~al.(2023{\natexlab{a}})Li, Li, Savarese, and Hoi]{blip2}
Junnan Li, Dongxu Li, Silvio Savarese, and Steven Hoi.
\newblock Blip-2: Bootstrapping language-image pre-training with frozen image encoders and large language models, 2023{\natexlab{a}}.

\bibitem[Li et~al.(2023{\natexlab{b}})Li, He, Wang, Li, Wang, Luo, Wang, Wang, and Qiao]{li2023videochat}
KunChang Li, Yinan He, Yi~Wang, Yizhuo Li, Wenhai Wang, Ping Luo, Yali Wang, Limin Wang, and Yu~Qiao.
\newblock Videochat: Chat-centric video understanding, 2023{\natexlab{b}}.

\bibitem[Li et~al.(2023{\natexlab{c}})Li, He, Wang, Li, Wang, Luo, Wang, Wang, and Qiao]{videochat}
KunChang Li, Yinan He, Yi~Wang, Yizhuo Li, Wenhai Wang, Ping Luo, Yali Wang, Limin Wang, and Yu~Qiao.
\newblock Videochat: Chat-centric video understanding.
\newblock \emph{arXiv preprint arXiv:2305.06355}, 2023{\natexlab{c}}.

\bibitem[Li et~al.(2019)Li, Yatskar, Yin, Hsieh, and Chang]{VisualBERT}
Liunian~Harold Li, Mark Yatskar, Da~Yin, Cho-Jui Hsieh, and Kai-Wei Chang.
\newblock Visualbert: A simple and performant baseline for vision and language, 2019.

\bibitem[Li* et~al.(2022)Li*, Zhang*, Zhang*, Yang, Li, Zhong, Wang, Yuan, Zhang, Hwang, Chang, and Gao]{GLIP}
Liunian~Harold Li*, Pengchuan Zhang*, Haotian Zhang*, Jianwei Yang, Chunyuan Li, Yiwu Zhong, Lijuan Wang, Lu~Yuan, Lei Zhang, Jenq-Neng Hwang, Kai-Wei Chang, and Jianfeng Gao.
\newblock Grounded language-image pre-training.
\newblock In \emph{CVPR}, 2022.

\bibitem[Li et~al.(2020)Li, Yin, Li, Hu, Zhang, Zhang, Wang, Hu, Dong, Wei, Choi, and Gao]{oscar}
Xiujun Li, Xi~Yin, Chunyuan Li, Xiaowei Hu, Pengchuan Zhang, Lei Zhang, Lijuan Wang, Houdong Hu, Li~Dong, Furu Wei, Yejin Choi, and Jianfeng Gao.
\newblock Oscar: Object-semantics aligned pre-training for vision-language tasks.
\newblock In \emph{ECCV}, 2020.

\bibitem[Liu et~al.(2023)Liu, Tao, Liu, Fan, Zhou, Huang, He, and Yang]{liu2023videocsr}
Tingkai Liu, Yunzhe Tao, Haogeng Liu, Qihang Fan, Ding Zhou, Huaibo Huang, Ran He, and Hongxia Yang.
\newblock Video-csr: Complex video digest creation for visual-language models, 2023.

\bibitem[Lu et~al.(2019)Lu, Batra, Parikh, and Lee]{vilbert}
Jiasen Lu, Dhruv Batra, Devi Parikh, and Stefan Lee.
\newblock Vilbert: Pretraining task-agnostic visiolinguistic representations for vision-and-language tasks.
\newblock In \emph{NeurIPS}, 2019.

\bibitem[Maaz et~al.(2023)Maaz, Rasheed, Khan, and Khan]{maaz2023videochatgpt}
Muhammad Maaz, Hanoona Rasheed, Salman Khan, and Fahad~Shahbaz Khan.
\newblock Video-chatgpt: Towards detailed video understanding via large vision and language models, 2023.

\bibitem[OpenAI(2023)]{openai2023gpt4}
OpenAI.
\newblock Gpt-4 technical report, 2023.

\bibitem[Radford et~al.(2021)Radford, Kim, Hallacy, Ramesh, et~al.]{CLIP}
Alec Radford, Jong~Wook Kim, Chris Hallacy, Aditya Ramesh, et~al.
\newblock Learning transferable visual models from natural language supervision.
\newblock In \emph{ICML}, 2021.

\bibitem[Rennie et~al.(2017)Rennie, Marcheret, Mroueh, Ross, and Goel]{rennie2017selfcritical}
Steven~J. Rennie, Etienne Marcheret, Youssef Mroueh, Jarret Ross, and Vaibhava Goel.
\newblock Self-critical sequence training for image captioning, 2017.

\bibitem[Schuhmann et~al.(2022)Schuhmann, Beaumont, Vencu, Gordon, Wightman, Cherti, Coombes, Katta, Mullis, Wortsman, Schramowski, Kundurthy, Crowson, Schmidt, Kaczmarczyk, and Jitsev]{schuhmann2022laion5b}
Christoph Schuhmann, Romain Beaumont, Richard Vencu, Cade Gordon, Ross Wightman, Mehdi Cherti, Theo Coombes, Aarush Katta, Clayton Mullis, Mitchell Wortsman, Patrick Schramowski, Srivatsa Kundurthy, Katherine Crowson, Ludwig Schmidt, Robert Kaczmarczyk, and Jenia Jitsev.
\newblock Laion-5b: An open large-scale dataset for training next generation image-text models, 2022.

\bibitem[Sellam et~al.(2020)Sellam, Das, and Parikh]{sellam2020bleurt}
Thibault Sellam, Dipanjan Das, and Ankur~P. Parikh.
\newblock Bleurt: Learning robust metrics for text generation, 2020.

\bibitem[Sharma et~al.(2018)Sharma, Ding, Goodman, and Soricut]{sharma2018conceptual}
Piyush Sharma, Nan Ding, Sebastian Goodman, and Radu Soricut.
\newblock Conceptual captions: A cleaned, hypernymed, image alt-text dataset for automatic image captioning.
\newblock In \emph{Proceedings of the 56th Annual Meeting of the Association for Computational Linguistics (Volume 1: Long Papers)}, pp.\  2556--2565, 2018.

\bibitem[Shukor et~al.(2022)Shukor, Couairon, and Cord]{shukor2022efficient}
Mustafa Shukor, Guillaume Couairon, and Matthieu Cord.
\newblock Efficient vision-language pretraining with visual concepts and hierarchical alignment.
\newblock \emph{arXiv preprint arXiv:2208.13628}, 2022.

\bibitem[Su et~al.(2020)Su, Zhu, Cao, Li, Lu, Wei, and Dai]{VL-BERT}
Weijie Su, Xizhou Zhu, Yue Cao, Bin Li, Lewei Lu, Furu Wei, and Jifeng Dai.
\newblock Vl-bert: Pre-training of generic visual-linguistic representations.
\newblock In \emph{ICLR}, 2020.

\bibitem[Tan \& Bansal(2019)Tan and Bansal]{lxmert}
Hao Tan and Mohit Bansal.
\newblock Lxmert: Learning cross-modality encoder representations from transformers.
\newblock In \emph{EMNLP}, 2019.

\bibitem[Touvron et~al.(2023)Touvron, Martin, Stone, Albert, et~al.]{touvron2023llama}
Hugo Touvron, Louis Martin, Kevin Stone, Peter Albert, et~al.
\newblock Llama 2: Open foundation and fine-tuned chat models, 2023.

\bibitem[Wang et~al.(2022{\natexlab{a}})Wang, Yang, Hu, Li, Lin, Gan, Liu, Liu, and Wang]{wang2022git}
Jianfeng Wang, Zhengyuan Yang, Xiaowei Hu, Linjie Li, Kevin Lin, Zhe Gan, Zicheng Liu, Ce~Liu, and Lijuan Wang.
\newblock Git: A generative image-to-text transformer for vision and language, 2022{\natexlab{a}}.

\bibitem[Wang et~al.(2022{\natexlab{b}})Wang, Yang, Men, Lin, et~al.]{wang2022ofa}
Peng Wang, An~Yang, Rui Men, Junyang Lin, et~al.
\newblock Ofa: Unifying architectures, tasks, and modalities through a simple sequence-to-sequence learning framework, 2022{\natexlab{b}}.

\bibitem[Xu et~al.(2023)Xu, Ye, Yan, Shi, Ye, Xu, Li, Bi, Qian, Wang, Xu, Zhang, Huang, Huang, and Zhou]{mplug2}
Haiyang Xu, Qinghao Ye, Ming Yan, Yaya Shi, Jiabo Ye, Yuanhong Xu, Chenliang Li, Bin Bi, Qi~Qian, Wei Wang, Guohai Xu, Ji~Zhang, Songfang Huang, Fei Huang, and Jingren Zhou.
\newblock mplug-2: A modularized multi-modal foundation model across text, image and video.
\newblock In \emph{ICML}, 2023.

\bibitem[Xu et~al.(2016)Xu, Mei, Yao, and Rui]{xu2016msr}
Jun Xu, Tao Mei, Ting Yao, and Yong Rui.
\newblock Msr-vtt: A large video description dataset for bridging video and language.
\newblock In \emph{Proceedings of the IEEE conference on computer vision and pattern recognition}, pp.\  5288--5296, 2016.

\bibitem[Yan et~al.(2023)Yan, Zhu, Wang, Cao, Zhang, Ghosh, Wu, and Yu]{videococa}
Shen Yan, Tao Zhu, Zirui Wang, Yuan Cao, Mi~Zhang, Soham Ghosh, Yonghui Wu, and Jiahui Yu.
\newblock Videococa: Video-text modeling with zero-shot transfer from contrastive captioners, 2023.

\bibitem[Yang et~al.(2022)Yang, Duan, Tran, Xu, Chanda, Chen, Zeng, Chilimbi, and Huang]{MoCo}
Jinyu Yang, Jiali Duan, Son Tran, Yi~Xu, Sampath Chanda, Liqun Chen, Belinda Zeng, Trishul Chilimbi, and Junzhou Huang.
\newblock Vision-language pre-training with triple contrastive learning.
\newblock In \emph{CVPR}, 2022.

\bibitem[Ye et~al.(2022)Ye, Xu, Yan, Xu, Qian, Zhang, and Huang]{ye2022hitea}
Qinghao Ye, Guohai Xu, Ming Yan, Haiyang Xu, Qi~Qian, Ji~Zhang, and Fei Huang.
\newblock Hitea: Hierarchical temporal-aware video-language pre-training, 2022.

\bibitem[Yu et~al.(2022)Yu, Wang, Vasudevan, Yeung, Seyedhosseini, and Wu]{coca}
Jiahui Yu, Zirui Wang, Vijay Vasudevan, Legg Yeung, Mojtaba Seyedhosseini, and Yonghui Wu.
\newblock Coca: Contrastive captioners are image-text foundation models, 2022.

\bibitem[Zeng et~al.(2022)Zeng, Zhang, and Li]{multigrained}
Yan Zeng, Xinsong Zhang, and Hang Li.
\newblock Multi-grained vision language pre-training: Aligning texts with visual concepts.
\newblock In \emph{CVPR}, 2022.

\bibitem[Zhang et~al.(2023{\natexlab{a}})Zhang, Fei, Yao, Ji, Li, Liu, and Chua]{zhang2023transfer}
Ao~Zhang, Hao Fei, Yuan Yao, Wei Ji, Li~Li, Zhiyuan Liu, and Tat-Seng Chua.
\newblock Transfer visual prompt generator across llms, 2023{\natexlab{a}}.

\bibitem[Zhang et~al.(2023{\natexlab{b}})Zhang, Li, and Bing]{videollama}
Hang Zhang, Xin Li, and Lidong Bing.
\newblock Video-llama: An instruction-tuned audio-visual language model for video understanding.
\newblock \emph{arXiv preprint arXiv:2306.02858}, 2023{\natexlab{b}}.

\bibitem[Zhang et~al.(2022)Zhang, Roller, Goyal, Artetxe, Chen, Chen, Dewan, Diab, Li, Lin, et~al.]{zhang2022opt}
Susan Zhang, Stephen Roller, Naman Goyal, Mikel Artetxe, Moya Chen, Shuohui Chen, Christopher Dewan, Mona Diab, Xian Li, Xi~Victoria Lin, et~al.
\newblock Opt: Open pre-trained transformer language models.
\newblock \emph{arXiv preprint arXiv:2205.01068}, 2022.

\bibitem[Zhu et~al.(2023)Zhu, Chen, Shen, Li, and Elhoseiny]{minigpt4}
Deyao Zhu, Jun Chen, Xiaoqian Shen, Xiang Li, and Mohamed Elhoseiny.
\newblock Minigpt-4: Enhancing vision-language understanding with advanced large language models.
\newblock \emph{arXiv preprint arXiv:2304.10592}, 2023.

\end{thebibliography}
\bibliographystyle{iclr2024_conference}

\newpage
\appendix
\section{Detailed algorithm description}
Three training stages are scheduled in our experiment. We will briefly show this with an algorithm table. Here we take video caption task as example. Where \ref{alg1} shows stage 1 while \ref{alg2} shows stage 2.
\begin{algorithm}[!h]
\label{alg1}
	%\textsl{}\setstretch{1.8}
	\renewcommand{\algorithmicrequire}{\textbf{Input:}}
	\renewcommand{\algorithmicensure}{\textbf{Output:}}
	\caption{Stage 1}
	\begin{algorithmic}[1]
        \STATE \textbf{Dataset}: 1.9M sentences, covering 0.5M summaries (long sentences) and 1.5M captions (short sentences). It should be noted that no testset's sentence are included.
	    \STATE \textbf{Model}: Text Auto-Encoder, mainly covering BERT and Text Q-Former (5 layers). About 150M trainable parameters, in which 110M are shared with the vision part.
        \STATE \textbf{Input}: $text$
        \REPEAT
        \STATE \textbf{Feature Decoupling}: $\mathbf{R}_{text}$ = $\mathbf{QF}_{text}$(BERT($text$)), where $\mathbf{QF}_{text}$ means Text Q-Former
        \STATE \textbf{Loss caculation}: $\mathcal{L}_{text} = \mathbf{LLM}(\mathbf{Proj}(\mathbf{R}_{text}), text).Loss$, where $\mathbf{Proj}$ and $\mathbf{LLM}$ means linear projection and large language model. 
        \STATE \textbf{Loss back-propagation and weights updating}
        \UNTIL convergence
	\end{algorithmic}  
\end{algorithm}

\begin{algorithm}[!h]
\label{alg2}
	%\textsl{}\setstretch{1.8}
	\renewcommand{\algorithmicrequire}{\textbf{Input:}}
	\renewcommand{\algorithmicensure}{\textbf{Output:}}
	\caption{Stage 2}
	\begin{algorithmic}[1]
        \STATE \textbf{Dataset}: WebVid2M, covering 2M valid (video, text) pairs; CC3M, covering 2.5M valid (image, text) pairs which is considered as static video.
	\STATE \textbf{Model}: Text Auto-Encoder, mainly covering BERT and Text Q-Former (5 layers). Video foundation model, mainly covering Image Q-Former, Video Q-Former (5 layers) and ASR BERT. Here ASR BERT are shared by both parts. About 307M trainable parameters, in which only 40M are exclusively belongs to text Auto-Encoder.
        \STATE \textbf{Input}: $(text, asr, frames)$, noting we use 'none.' to represent the videos' asr for the pre-training dataset.
        \REPEAT
        \STATE  \# A. Text Auto-Encoder pipeline.
        \STATE \textbf{Feature Decoupling}: $\mathbf{R}_{text}$ = $\mathbf{QF}_{text}$(BERT($text$)), where $\mathbf{QF}_{text}$ means Text Q-Former
        \STATE \textbf{Loss caculation}: $\mathcal{L}_{text} = \mathbf{LLM}(\mathbf{Proj}(\mathbf{R}_{text}), text).Loss$, where $\mathbf{Proj}$ and $\mathbf{LLM}$ means linear projection and large language model. 
        \STATE
        \STATE  \# B. Video pipeline.
        \STATE \textbf{ASR processing}: $\mathbf{R}_{asr}$ = BERT($asr$), 
        \STATE \textbf{Video frames processing}: $\mathbf{R}_{vision}$ = $\mathbf{QF}_{img}$(ViT($frames$)), where $\mathbf{QF}_{img}$ means iamge Q-Former in BLIP-2.
        \STATE \textbf{Video feature decoupling}: $\mathbf{R}_{video}$ = $\mathbf{QF}_{video}$(Concat($\mathbf{R}_{vision},\mathbf{R}_{asr}$))
        \STATE \textbf{Loss caculation}: $\mathcal{L}_{contra}$ is just caculated by the first token from $\mathbf{R}_{video}, \mathbf{R}_{text}$
         \[
        \begin{aligned}
            \mathcal{L}_{video} = \mathbf{LLM}(\mathbf{Proj}(\mathbf{R}_{video}), text).Loss
        \end{aligned}
        \nonumber
        \]
        \[
        \begin{aligned}
            \mathcal{L}_{align} = \mathrm{MSE}(\mathbf{R}_{video}, \mathbf{R}_{text})
        \end{aligned}
        \nonumber
        \]
        \STATE  \# C. Loss combination.
        \STATE \textbf{Total training loss}: $\mathcal{L}_{train}=\mathcal{L}_{video}+\mathcal{L}_{contra}+\lambda_1 \mathcal{L}_{text}+\lambda_2\mathcal{L}_{align}$  
        \STATE \textbf{Loss back-propagation and weights updating}
        \UNTIL convergence
	\end{algorithmic}  
\end{algorithm}
The training stage three is the same with stage \ref{alg2}, where the only difference is the dataset.

\section{Demos}
\begin{figure}[H]
    \centering
    \includegraphics[width=0.85\linewidth]{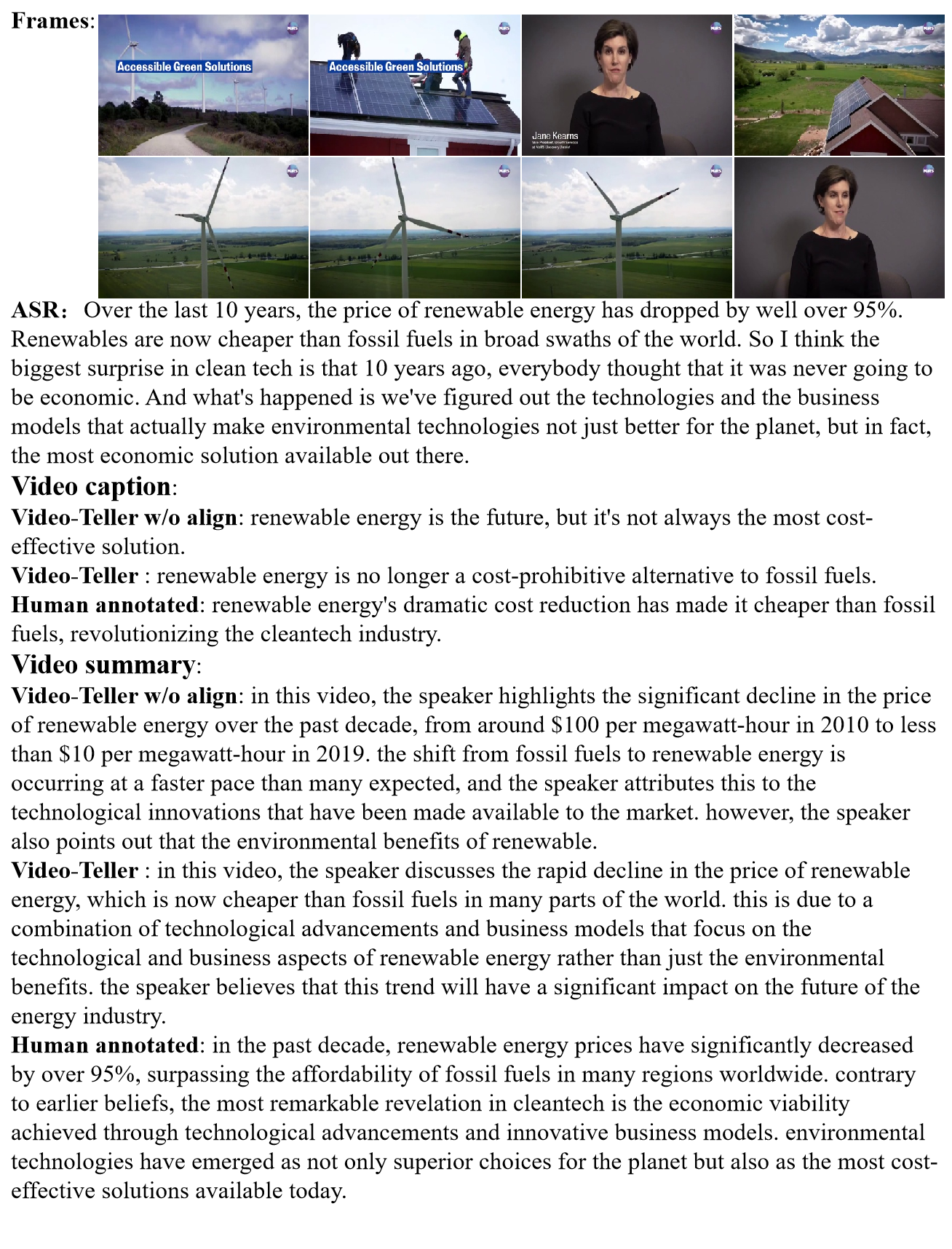}
    \caption{A case shows improvement of hallucination using fine-grained modality alignment.}
    \label{fig:2}
\end{figure}


\end{document}